\documentclass[lettersize,journal]{IEEEtran}
\usepackage{amsmath,amsfonts}

\usepackage{algorithmic}
\usepackage{algorithm}
\usepackage{setspace}
\usepackage{array}
\usepackage[caption=false,font=normalsize,labelfont=sf,textfont=sf]{subfig}
\usepackage{textcomp}
\usepackage{stfloats}
\usepackage{url}
\usepackage{tablefootnote}
\usepackage{threeparttable}
\usepackage{graphicx} 
\usepackage{siunitx}
\usepackage{verbatim}
\usepackage{graphicx}
\usepackage{cite}
\usepackage{multirow}
\usepackage[inline]{enumitem}

\usepackage{todonotes}
\newcommand{\vtheta}{\ensuremath{\mathrm{\boldsymbol{\theta}}}}

\newcommand{\vy}{\ensuremath{\mathrm{\bold{y}}}}
\newcommand{\vb}{\ensuremath{\mathrm{\bold{b}}}}

\newcommand{\mM}{\ensuremath{\mathrm{\bold{M}}}}
\newcommand{\mW}{\ensuremath{\mathrm{\bold{W}}}}
\allowdisplaybreaks

\newcommand{\x}{\mathbf{x}}

\newcommand{\y}{\mathbf{y}}

\newcommand{\Z}{\mathbf{Z}}

\newcommand{\W}{\mathbf{W}}

\def\fl[#1\]{\begin{align}#1\end{align}}
\def\[#1\]{\begin{align*}#1\end{align*}}

\def\*[#1\]{\begin{align*}#1\end{align*}}
\begin{document}
\bstctlcite{IEEEexample:BSTcontrol}

\title{\huge Spatial-SpinDrop: Spatial Dropout-based Binary Bayesian Neural Network with Spintronics Implementation}

\author{\IEEEauthorblockN{Soyed Tuhin Ahmed\IEEEauthorrefmark{9}\IEEEauthorrefmark{2}, Kamal Danouchi\IEEEauthorrefmark{3}, Michael Hefenbrock\IEEEauthorrefmark{4}, Guillaume Prenat\IEEEauthorrefmark{3}, Lorena Anghel\IEEEauthorrefmark{3}, Mehdi B. Tahoori\IEEEauthorrefmark{2}\\}
\IEEEauthorblockA{\IEEEauthorrefmark{2}Karlsruhe Institute of Technology, Karlsruhe, Germany, \IEEEauthorrefmark{9}corresponding author, email: soyed.ahmed@kit.edu}
\IEEEauthorblockA{\IEEEauthorrefmark{3}Univ. Grenoble Alpes, CEA, CNRS, Grenoble INP, and IRIG-Spintec, Grenoble, France}
\IEEEauthorblockA{\IEEEauthorrefmark{4}RevoAI GmbH, Karlsruhe, Germany}}



\maketitle

\begin{abstract}
Recently, machine learning systems have gained prominence in real-time, critical decision-making domains, such as autonomous driving and industrial automation. Their implementations should avoid overconfident predictions through uncertainty estimation. Bayesian Neural Networks (BayNNs) are principled methods for estimating predictive uncertainty. However, their computational costs and power consumption hinder their widespread deployment in edge AI. Utilizing Dropout as an approximation of the posterior distribution, binarizing the parameters of BayNNs, and further to that implementing them in spintronics-based computation-in-memory (CiM) hardware arrays provide can be a viable solution. However, designing hardware Dropout modules for convolutional neural network (CNN) topologies is challenging and expensive, as they may require numerous Dropout modules and need to use spatial information to drop certain elements. In this paper, we introduce MC-SpatialDropout, a spatial dropout-based approximate BayNNs with spintronics emerging devices. Our method utilizes the inherent stochasticity of spintronic devices for efficient implementation of the spatial dropout module compared to existing implementations. Furthermore, the number of dropout modules per network layer is reduced by a factor of $9\times$ and energy consumption by a factor of $94.11\times$, while still achieving comparable predictive performance and uncertainty estimates compared to related works.
\end{abstract}


\addtolength\abovedisplayskip{-0.4em}%
\addtolength\belowdisplayskip{-0.4em}%
\setlength{\textfloatsep}{0.5pt}
\setlength\belowdisplayskip{0pt}

\section{Introduction}
\IEEEPARstart{N}{eural networks} are brain-inspired computational methods that, in some cases, can even outperform human counterparts~\cite{kiela2021dynabench}. Consequently, applications of NNs have increased rapidly in recent years and have become the cornerstone of modern computing paradigms. Furthermore, NNs are commonly deployed in real-time safety-critical tasks such as computer-aided medical diagnostics, industrial robotics, and autonomous vehicles. 
% However, there are challenges when using NNs in safety-critical applications that must be addressed to assure trust in the system.

Conventional (point estimate) Neural Networks (NNs) typically learn a single point value for each parameter. However, they do not account for the uncertainty in the data nor in the model, leading to overconfident predictions and in turnn to safety violations. This is particularly true when the data generation process is noisy or the training data is either incomplete or insufficient to capture the complexity of the actual phenomenon being modelled. In safety-critical domains where machine learning systems make human-centered decisions, an uncertainty measure is essential for informed decision-making.

On the other hand, Bayesian Neural Networks (BayNNs), which put prior distributions over the model parameters and learn the posterior distribution using approximation techniques (e.g., Monte Carlo (MC)-Dropout~\cite{gal2016dropout}), present a systematic method for training uncertainty-aware neural networks. However, the computational costs and high-performance requirements of BayNNs can be prohibitive for edge devices. 
% The inherent computational complexity and performance demands of BayNNs can pose challenges for deployment on embedded devices.

Therefore, dedicated NN hardware accelerators such as Compute-in-Memory (CiM) architectures with emerging Non-Volatile resistive Memories (NVMs) have been explored. CiM architectures enable the Matrix-Vector Multiplication (MVM) operation of NNs to be carried out directly inside the memory, overcoming the memory limitations of traditional von-Neumann architectures. Among the NVM technologies, Spin-Transfer-Torque Magnetic Random Access Memory (STT-MRAM) is particularly appealing due to its nanosecond latency, high endurance ($10^{12}$ cycles), and low switching energy (10 fJ)~\cite{wang2020resistive}.

Additionally, algorithmic approaches, such as Binarization which typically reduces the bit precision of NNs to 1-bit, lead to smaller computational time and model size. Therefore, they are an attractive options for BayNNs to mitigate their inherent costs. Moreover, this approach allows for the direct mapping of BayNN parameters to STT-MRAM-based CiM hardware.

% The preliminary version of this 
Existing work~\cite{soyed_nanoarch22,soyed_spindrop} proposed to binarize the parameters of BayNNs and implement them on STT-MRAM-based CiM hardware resulting in a highly efficient solution. Although this approach can achieve high algorithmic performance and hardware efficiency compared to existing works, designing Dropout modules in the case of convolutional NN (CNN) topologies is challenging and expensive due to the nature of implementation.

In this paper, we present an algorithm-hardware co-design approach that not only solves the challenges of implementing the Dropout-based BayNNs approach, but also reduces the number of Dropout modules required per layer. The main contributions of this paper are as follows:

\begin{itemize}
    \item We propose \emph{MC-SpatialDropout}, which uses spatial Dropout for Bayesian approximation. Our method is mathematically equivalent to the MC-Dropout-based approach, enabling uncertainty-aware predictions. % similar to the learning objective of Gaussian processes.
    % \item We introduce an uncertainty estimation metric capable of detecting out-of-distribution and noisy data with high accuracy, and demonstrate improved prediction accuracy when a subset of the test data is corrupted.
    \item We present an STT-MRAM-based CiM architecture for the proposed \emph{MC-SpatialDropout}-based BayNNs. Our approach leverages the inherent stochasticity of STT-MRAM for the Dropout module and deterministic behavior for parameter storage. This allows the reuse of the array designed for conventional binary NNs (BNNs), and only the peripheral circuitry is adapted for Bayesian inference.
    \item We also propose reliable and adaptable sensing scheme for stochastic STT-MRAM specifically designed to implement the dropout concept for both linear and convolutional layers.
\end{itemize}

% combines the benefits of Bayesian neural networks, binarization, and specialized hardware accelerators. We introduce a Bayesian binary neural network (BayBNN) using dropout-based approximation and leverage the inherent stochastic properties of STT-MRAM for in-memory Bayesian inference. Furthermore, we 
Our method is targeting CNN topologies and reduces the number of Dropout modules in a layer by $9\times$ and energy consumption by $94.11\times$, while maintaining comparable predictive performance and uncertainty estimates.

The remainder of this paper is organized as follows: Section \ref{sec:background} provides the background for our work, Section~\ref{sec:baynn_dropout} describes the proposed MC-SpatialDropout, Section \ref{sec:Results} presents both the algorithmic and hardware results for our approach and finally, in Section~\ref{sec:conclusion}, we conclude the paper.

\section{Background}\label{sec:background}
\subsection{Spintronics}
MRAM have gained significant attention due to their fast switching, high endurance, and CMOS compatibility \cite{lee_world-most_2022}. The main component of MRAM devices is the Magnetic Tunnel Junction (MTJ), which comprises two ferromagnetic layers: the reference layer and the free layer, separated by a thin insulating layer. The magnetization of the reference layer is fixed in one direction, while the free layer can have its magnetization reversed between two stable positions: parallel or antiparallel to that of the reference layer. The resistance of the stack depends on the relative orientations of the layer magnetizations, with a high resistance state in the antiparallel configuration and a low resistance state in the parallel configuration.
\vspace{-0.5\baselineskip}

% \subsection{Dropout}

% \todo[inline]{I don't quite get this part. From my understanding, a feature map ( shape ($H \times W \times 1$) ) is the result of the computation of one kernel. And for a conv layer, a kernel is essentially a neuron, i.e., a weighted sum of inputs, possibly with activation. In this case, "neuron (kernel)" dropout is exactly the same as neuron dropout (setting the output of the neuron/kernel =0 for all its computations). Maybe we could say that normal dropout operates on the individually calculated activations, i.e., activation of neuron in MLP/ entries of feature map in CNN, and spacial dropout operates on the entire feature map at once in CNNs.
% Hence, (activation) dropout $\mM_{i,j,k} \sim \mathcal{B}(p)$ vs spacial dropout $\mM = \mathbf{1} \otimes \mathbf{1} \otimes \vm$ with $\vm_k  \sim B(p)$. In both cases, we apply $\mM$ to an activation tensor, so $\va \leftarrow \mM \odot \va$}
% Spatial dropout is a variant of conventional dropout that aims to improve the efficiency of convolutional neural networks. In spatial dropout, entire feature maps are dropped out rather than individual neurons. The mathematical equation for spatial dropout is similar to conventional dropout, but the binary mask $\boldsymbol{M}$ is applied to the entire feature map.

\subsection{Uncertainty in Deep Learning}
Uncertainty estimation is vital in deep learning, especially for safety-critical applications, as it provides insight into the model's confidence in its predictions, enhancing the trustworthiness of decision-making. There are two main types of uncertainty: epistemic, which results from the limitations of the model and can be reduced with more data or improved architectures, and aleatoric, which arises from noise in the data and cannot be mitigated. Obtaining uncertainty estimates bolsters robustness by identifying out-of-distribution (OOD) data points and avoiding overconfident predictions. OOD data refers to data whose distribution is completely different from the training (in-distribution (ID)) data. In this paper, we focus on aleatoric uncertainty estimation and evaluate the effectiveness of our method for OOD detection.

\vspace{-0.5\baselineskip}
\subsection{Bayesian NNs}
BayNNs offer a principled approach to uncertainty estimation in neural networks. Several approximation methods exist for BayNNs, such as variational inference and Markov Chain Monte Carlo methods.

% \todo[inline]{Maybe we should not overload x and y here for the definition of the mask. Also, $\mM$ is a matrix (since we used an uppercase latter) while x and y, use the lowercase (vector) notation. If we take our notation wrt to this serious, $\boldsymbol{M} \odot \boldsymbol{x}$ would not work. Maybe define it }
One popular approximation technique is Monte Carlo Dropout (MC-Dropout), which leverages dropout for Bayesian inference. Dropout \cite{srivastava2014dropout} is a common regularization technique used to reduce overfitting and neuron co-adaptation by randomly setting neuron outputs to zero during training. The dropout operation can be described as $\hat{\Z} = \mM \odot \Z$, where $\mM$ is a binary mask generated by sampling from a Bernoulli distribution, $\odot$ represents element-wise multiplication, and $\Z$ and $\hat{\Z}$ are intermediate activation and dropped out intermediate activation of a layer, respectively.

MC-Dropout provides an approximation of the true posterior distribution with relatively low computational and memory overhead compared to other methods such as variational inference (VI)~\cite{blundell2015weight} and the ensemble approach~\cite{lakshminarayanan2017simple}. 
This is because the ensemble approach requires inference in multiple NNs, and VI requires learning the parameters of the variational distribution, which require storage. Since the MC-Dropout method has the same number of parameters as conventional NNs, it leads to minimal additional computation and memory requirements, making it suitable for a wide range of applications, including those with limited resources.

The optimization objective for MC-Dropout can be represented as 
% \todo[inline]{Look at the latex code of this comment since there seems to be some confusion: if you use the symbols I defined, like $\vtheta$ you don't need to use "boldsymbol" round it (like $\boldsymbol{\vtheta}$). Same with $\vx, \vy, \mM, \mW$ etc.}
% \todo[inline]{Are you sure that the original MC dropout regularizes all parameters? Including the bias ? As you are including $\vtheta$ here.}
\begin{equation}\label{eq:MC_dropout}
\mathcal{L}(\boldsymbol{\vtheta})_{\text{MC-Dropout}} = \mathcal{L}(\vtheta, \mathcal{D}) + \lambda\sum_{l=1}^L  ||\vtheta_l||^2_2
\end{equation}
% \todo[inline]{Why don't you just call the (data) loss $\mathcal{L}(\vtheta)$ or $\mathcal{L}(\vtheta, \mathcal{D})$. I don't think we need the c. In the case of regression you mention, there are no classes and the formulation would then be confusing. Additionally, the $\hat{y}_c$ and $y_c$ are not defined and not connected to \eqref{eq:bayesian} or \eqref{eq:mc_dropout_posteior_approx}.}
% \todo[inline]{If you write $\Vert \vtheta \Vert$, do you also mean to regularize the biases? Because you define $\vtheta$ as all the learnable parameters it seems. Later you also specifically refer to the weights separately. The relationship between $\vtheta$ and the weights should be made more clear. Possibly state that $\vtheta$ summarizes all learnable parameters, i.e., $\vtheta = \{ \mW^l, \vb^l \mid l =1, \cdots, L\}$, where $\mW^l$ denote the weight matrices and $\vb^l$ the biases for the layer $l$.}
where $\mathcal{L}(\vtheta, \mathcal{D})$ represents the task-specific loss function, such as categorical cross-entropy for classification or mean squared error for regression, and $\|\boldsymbol{\vtheta}\|^2_2$ is the regularization term. Also, $\boldsymbol{\vtheta}$ summarizes all learnable parameters, i.e., $\vtheta = \{\mW_l, \vb_l \mid l =1, \cdots, L\}$, where $\mW_{l}$ denote the weight matrices and $\vb_l$ the biases for the layer $l$. 
During inference, dropout is applied multiple times, and the outputs are averaged to obtain the predictive distribution. Hence, the posterior predictive distribution over the output $y$, i.e.,
\begin{equation}\label{eq:bayesian}
p(\vy \vert \x, \mathcal{D}) = \int p(\vy|\x, \boldsymbol{\vtheta}) p(\boldsymbol{\vtheta}|\mathcal{D}) d\boldsymbol{\vtheta}
\end{equation}
is approximated by 
% \todo[inline]{When using dropout, the parameters $\theta$ are not random variables. Only the dropout masks are random.}
\begin{equation} \label{eq:mc_dropout_posteior_approx}
p(\y|\x, \mathcal{D}) \approx \frac{1}{T} \sum_{t=1}^T p(\y|\x, \vtheta, \mM_t) \quad \text{with} \quad \mM_t \sim \mathcal{B}(\rho).
\end{equation}
% \todo[inline]{When using dropout, the parameters $\theta$ are not random variables. Only the dropout masks are random.}
Here, $\mathcal{D}$ denotes the dataset, $\x$ is the input, $\y$ is the output, and the entries of $\mM_t$ are independently sampled from a Bernoulli distribution with (dropout) probability $\rho$.

% During inference, dropout is applied multiple times in each of the $T$ forward passes, and the outputs, $y_1\cdots y_T$, are averaged to obtain the posterior probability. The Bayesian inference in MC-Dropout can be formulated as 
% \begin{equation}
% p(y^*|\x^*, \mathcal{D}) \approx \frac{1}{T} \sum_{t=1}^T p(y^*|\x^*, \boldsymbol{W}_t).
% \end{equation}
% Where $\mathcal{D}$ is the dataset, $\x^*$ is the input, $y^*$ is the output, and $\boldsymbol{W}_t$ represents the sampled weights at the $t$-th forward pass. 
% \begin{equation}
% p(y^*|\x^*, \mathcal{D}) \approx \frac{1}{T} \sum_{t=1}^T p(y^*|\x^*, \vtheta, \mM_t) \quad \mM_t \sim \mathcal{B}(p).
% \end{equation}
% Where $\mathcal{D}$ is the dataset, $\x^*$ is the input, $y^*$ is the output, and $\mM_t$ sampled element-wise from a Bernoulli distribution. 
\vspace{-0.5\baselineskip}

\subsection{Mapping of Convolutional Layers to CiM Architecture}\label{sec:mapping}
To perform the computation inside the CiM architecture, a critical step is the mapping of the different layers of the NN to crossbar arrays. Standard NNs contain mainly Fully Connected (FC) layers and convolutional layers. While the mapping of FC layers is straightforward in a crossbar array as the shape of the weight matrices is 2D ($\mathbb{R}^{m \times n}$), mapping convolutional layers is challenging due to their 4D shapes ($\mathbb{R}^{K \times K \times C_{in} \times C_{out}}$). Here, $K$ denotes the shape of kernels, and $C_{in}$ represents the number of input channels. Implementing convolutional layers requires implementing multiple kernels with different shapes and sizes. 

There are two popular mapping strategies for mapping the convolutional layer exists. In the mapping strategy \textcircled{1}, each kernel of shape $K\times K\times C_{in}$ is unrolled to a column of the crossbar~\cite{gokmen_training_2017}. On the other hand, in the mapping strategy \textcircled{2}, each kernel is mapped to $K\times K$ smaller crossbars with a shape of $C_{in}\times C_{out}$~\cite{peng_optimizing_2019}.
% We refer to crossbar implementing convolutional layer as conv-crossbar.

% Furthermore, implementing dropout with the method proposed in \cite{soyed_nanoarch22, soyed_spindrop} might not be efficient when considering convolutional layers. The scheme proposed in their study requires implementing one dropout module per neuron, which can become power and area inefficient as the size of the neural network increases.

% \subsection{Uncertainty Estimates in Deep Learning}
% Uncertainty estimation plays a crucial role in deep learning applications, particularly in safety-critical domains. It provides a measure of the model's confidence in its predictions, allowing for more reliable and informed decision-making.

% There are two primary types of uncertainty: epistemic and aleatoric. Epistemic uncertainty arises from the model's lack of knowledge, which can be reduced with more data or better model architectures. Aleatoric uncertainty stems from inherent noise in the data, and it cannot be reduced even with more data.

% In deep learning, obtaining uncertainty estimates is essential. Uncertainty estimates can improve robustness by identify out-of-distribution data points, allowing the user to react accordingly and avoid making overconfident predictions. In safety-critical applications, having uncertainty estimates can aid in making more informed decisions by understanding the confidence level of the model's predictions.
\vspace{-0.5\baselineskip}
\section{Proposed Method}\label{sec:baynn_dropout}
\subsection{Problem Statement and Motivation}

\begin{figure}
    \centering
    \includegraphics[width=0.85\columnwidth]{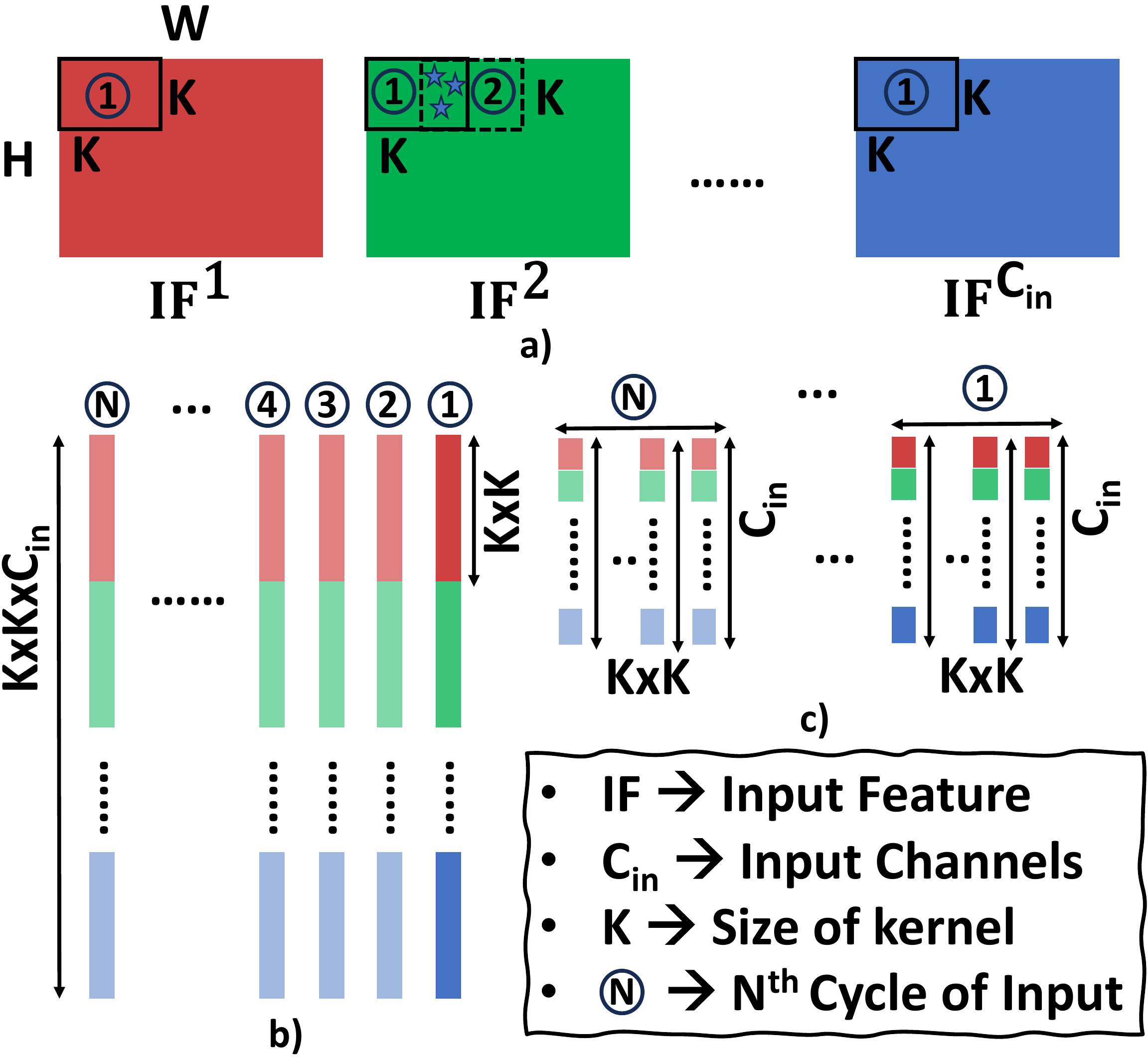}
    \caption{a) Input feature map of a convolutional layer, b) moving windows from all the input feature maps are flattened for the conventional mapping, c)  }
    \label{fig:mapping_MWs}
\end{figure}

The convolution operation is performed differently in CiM architectures compared to GPUs. In CiM architectures, moving windows (MWs) with a shape of $K\times K$ are applied to each input feature map (IFM) in one cycle (see Fig.~\ref{fig:mapping_MWs}(a)). In the next cycle, the MWs will "slide over" the IFMs with a topology-defined stride $S$ for $N$ cycles. Assuming $K>S$, some of the elements in the MWs for the next $K-S$ cycles will be the same as in the previous cycles, a concept known as weight sharing. This is illustrated by the green input feature (IF) in Fig.~\ref{fig:mapping_MWs}(a).

% In conventional dropout, elementwise dropout operation is applied to the IFMs of a layer is dropped. Unlike normal Dropout, Spatial Dropout discards entire feature maps instead of dropping individual elements.
%In conventional Dropout, each of the elements IFMs are dropped with a probability $P$ but dropped elements remain consistent throughout the $N$ cycles. However, t
The Dropout module designed in~\cite{soyed_nanoarch22,soyed_spindrop} drops each element of the MWs with a probability $P$ in each cycle. Therefore, it essentially re-samples the dropout mask of each MW of IFMs in each cycle. Consequently, the dropout masks of the shared elements in the MWs will change in each input cycle, leading to inconsistency. An ideal Dropout module should only generate dropout masks for new elements of the MWs. Designing a Dropout module that drops each element of the MWs depending on the spatial location of the MWs in the IFMs is challenging and may lead to complex circuit design. Additionally, the number of rows in crossbars typically increases from one layer to another due to the larger $C_{in}$. Consequently, the number of Dropout modules required will be significantly higher.

% Therefore, in this paper, we propose MC-SpatialDropout based 

Furthermore, the MWs are reshaped depending on the weight mapping discussed in Section~\ref{sec:mapping}. For mapping strategy \textcircled{1}, the MWs from IFMs are flattened into a vector of length $K\times K\times C_{in}$. However, for mapping strategy \textcircled{2}, IFMs are flattened into $K\times K$ vectors of length $C_{in}$, as depicted in Fig.~\ref{fig:mapping_MWs}(a) and (b). As a result, designing a generalizable Dropout model is challenging. 
% Our proposed MC-SpatialDropout, with a spintronic-based implementation, considers different mapping strategies and locations where Dropout is applied, offering a generalizable approach.

\vspace{-.5\baselineskip}
\subsection{MC-SpatialDropout as Bayesian Approximation}
In an effort to improve the efficiency and accuracy of Bayesian approximation techniques, we propose the MC-SpatialDropout method. The proposed MC-SpatialDropout technique expands upon the MC-Dropout~\cite{gal2016dropout} and MC-SpinDrop~\cite{soyed_nanoarch22,soyed_spindrop} methods by utilizing spatial dropout as a Bayesian approximation. Our approach drops an entire feature with a probability $p$. This means that all the elements of a feature map in Fig.~\ref{fig:mapping_MWs}(a) are dropped together. However, each feature map is dropped independently of the others. As a result, the number of Dropout modules required for a layer will be significantly reduced, and the design effort of the dropout module will also be lessened.

The primary objective of this approach is to address the shortcomings of MC-Dropout arising from its independent treatment of elements of the features. In contrast, MC-SpatialDropout exploits the spatial correlation of IFs, which is particularly advantageous for tasks involving image or spatial data. By doing so, it facilitates a more robust and contextually accurate approximation of the posterior distribution. This enables the model to capture more sophisticated representations and account for dependencies between features.

% showed that minimizing the L2 regularization term is mathematically equivalent to minimizing the Kullback–Leibler (KL) divergence term for Bayesian neural networks (NNs). 
In terms of the objective function for the MC-SpatialDropout, Soyed et al.~\cite{soyed_nanoarch22,soyed_spindrop} showed that minimizing the objective function of MC-Dropout (see Equation~\eqref{eq:MC_dropout}) is not beneficial for BNNs and suggested a BNN-specific regularization term. In this paper, instead of defining a separate loss function for MC-SpatialDropout, we define the objective function as:
% \todo[inline]{The right side of this equation is exactly the same as equation~\eqref{eq:MC_dropout}.}
\begin{equation} \label{eq:MC-SpatialDropout}
\mathcal{L}({\vtheta})_{\text{MC-SpatialDropout}} = \mathcal{L}(\vtheta, \mathcal{D}) + \lambda\sum_{l=1}^L  ||\mW_l||^2_2 .
\end{equation}
% Here, the first part $\frac{1}{C} \sum_{c=1}^C E(\y_c,\hat{\y}_c)$ represents the task-specific loss function, such as categorical cross-entropy for classification or mean squared error for regression, where $C$ denotes the number of classes in the dataset. 
Therefore, the objective function is equivalent to Equation~\eqref{eq:MC_dropout} for MC-Dropout. However,
the second part of the objective function is the regularization term applied to the (real valued) "proxy" weights ($\mW_l$) of BNN instead of binary weights. It encourages $\mW_l$ to be close to zero. By keeping a small value for the $\lambda$, it implicitly ensures that the distribution of weights is centered around zero.
Also, we normalize the weights by
\begin{equation}
    \bar{\mW}_l = \frac{\mW_l-\mu^{\mW}_l}{\sigma^{\mW}_l},
\end{equation}
to ensure, the weight matrix has zero mean and unit variance before binarization. Where $\mu^{\W}$ and $\sigma^{\W}$ are the mean and variance of the weight matrix of the layer $l$. This process allows applying L2 regularization in BNN training and~\cite{irnet} showed that it improves inference accuracy by reducing quantization error. Since our work is targeted for BNN, regularization is only applied to the weight matrixes.

% This allows us to take advantage of the spatial correlation between features when performing tasks involving spatial or image data.

The difference is that our method approximate Equation~\eqref{eq:bayesian} by:
% \todo[inline]{Look at this in Latex: To refer to equations, you can use Equation~\eqref{eq:bayesian} instead of Equation~\ref{eq:bayesian}}
\begin{equation} \label{eq:mc_spatialdropout_posteior_approx}
p(\y|\x, \mathcal{D}) \approx \frac{1}{T} \sum_{t=1}^T p(\y|\x, \vtheta, \hat{\mM}_t) \quad \text{with} \quad \hat{\mM}_t \sim \mathcal{B}(\rho).
\end{equation}
% \todo[inline]{This equation also looks exactly like the equation before besides the small hat on M. Maybe we could just define how we sample M and then refer to the previous equation.}
% Where, the mask $\hat{\mM}_t$ is sampled  instead of
Here, during training and Bayesian inference, the dropout mask $\hat{\mM}_t$ sampled spatially correlated manner for the output feature maps (OFMs) of each layer
% \emph{spatial} element-wise 
from a Bernoulli distribution with (dropout) probability $\rho$. 
% That means, we in . 
The dropout masks correspond to whether a certain spatial location in the OFMs (i.e., a certain unit) is dropped or not. 

For Bayesian inference, we perform $T$ Monte Carlo sampling to approximate the posterior distribution. Each Monte Carlo sample corresponds to forward passing the input $\x$ through the NN with unique spatial dropout masks $\hat{\mM}_t \quad t=1, \cdots, T$, resulting in a diverse ensemble of networks. By averaging the predictions from the Monte Carlo samples, we effectively perform Bayesian model averaging to obtain the final prediction.
% For inference with MC-SpatialDropout, we utilize Monte Carlo (MC) sampling. This method effectively approximates the posterior distribution, providing reliable uncertainty quantification. Each MC sample corresponds to a different dropout mask, leading to a diverse ensemble of networks. By performing MC sampling, we essentially generate a multitude of plausible models given the data, each contributing to the final prediction. This ensemble approach mitigates the risk of overfitting to the training data, ensuring the robustness and generalizability of the model.

% \begin{equation}
%     p(\y|\x) \approx \frac{1}{T} \sum_{t=1}^{T} p(\y|\x,\W_t).
% \end{equation}
% \todo[inline]{Why is $\mathbf{w}$ small here, while above it is big? Is it a vector or a matrix?}
% \todo[inline]{Is this not the same as Equation~\eqref{eq:mc_dropout_posteior_approx}? Just that M is sampled slightly differently?}
% \todo[inline]{Please clean up the notation. I don't think we need the $*$ for the $x^*$ because we always write $\mathcal{D}$ when we refer to the data. So i would suggest simply to use $\vx$ everywhere.}
% \todo[inline]{Here you use $\vx$ and $\vy$ while in the Background section on Bayesian NN you used only $x$ and $y$ (or with $*$ respectively). Decide for one of them and make it consistent. I would go with $\vx$ and $\vy$ to indicate that they are vectors.}

% Where $\y$ is the final output, $\mathbf{x^*}$ is the input, and $\mathbf{w}_t$ denotes the weights of the model at the $t$-th dropout mask. 

\begin{figure}
    \centering
    \includegraphics[width=0.9\columnwidth]{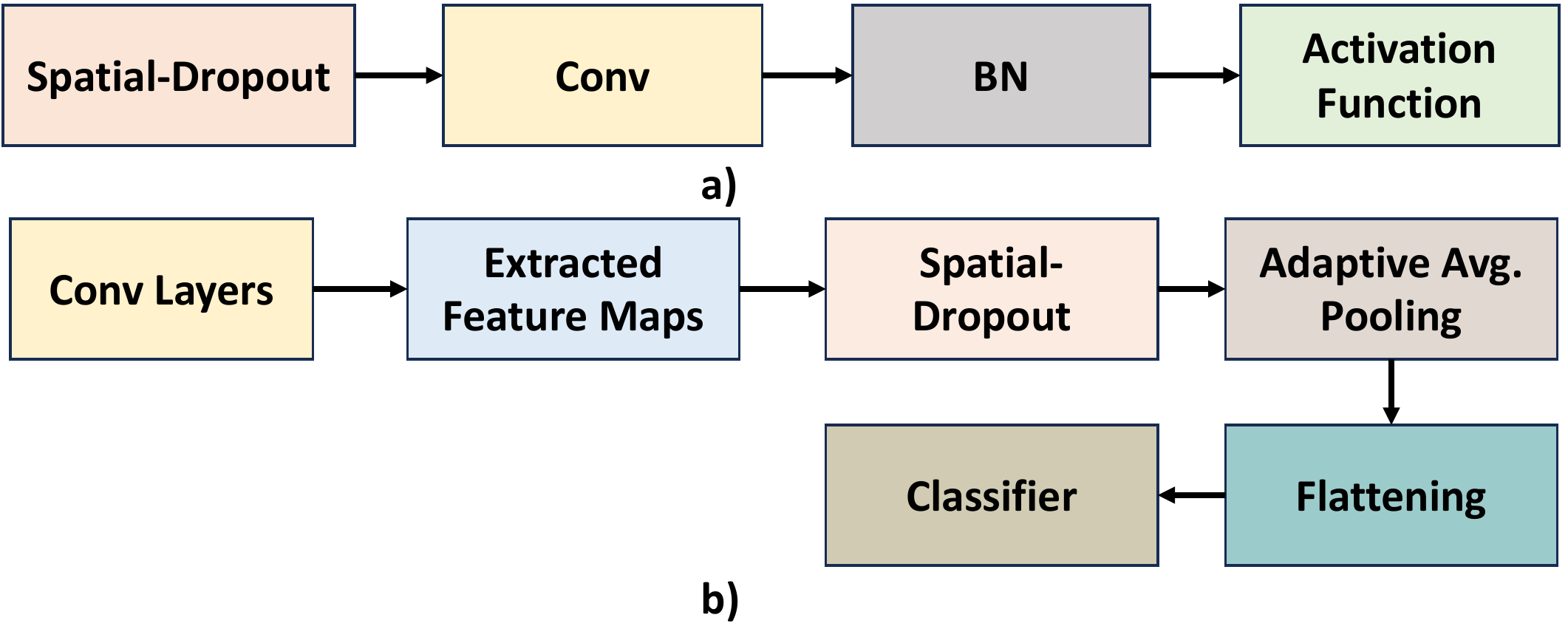}
    \caption{Block diagram of the location of the proposed MC-SpatialDropout in a) a layer-wise fashion, b) a topology specific fashion.}
    \label{fig:block_drop}
\end{figure}

Proper arrangement of layers is important for the MC-SpatialDropout based Bayesian inference. The Spatial Dropout layer can be applied before each convolutional layer in a layerwise MC-SpatialDropout method. Additionally, the Spatial Dropout layer can be applied to the extracted features of a CNN topology in a topology-wise MC-SpatialDropout method. Fig.~\ref{fig:block_drop} shows the block diagram for both approaches. 
% The block diagram for both approaches is depicted in Fig.~\ref{fig:block_drop}.

%\subsection{Mathematical Modelling of Spatial %Dropout Implemented with Spintronics}
%\label{sec:Mathematical Modelling of Spatial Dropout Implemented with Spintronics}
%In presence of the thermal noise, the switching of the MTJ has a stochastic behavior: for a given  amplitude and duration of a current passing through the MTJ, the switching has a certain probability to occur, expressed as: 
%\begin{equation}
%P(I) = 1 - e^{\frac{t}{\tau}}
%\end{equation}
%With $\tau = \tau_0e^{-\Delta\left(1-\frac{I}{I_{c0}}\right)^2}$
%Where $\Delta$ is the thermal stability factor, $t$ is the pulse duration, $\tau_0$ is the attempt time and $I_{c0}$ is the critical current at 0$K$ temperature. 
%Fig.~\ref{fig:MC-STT}, presents the switching probability of an MTJ with respect to the applied voltage. From left to right, the curves show the impact on the switching probability when the voltage level is decreased \cite{vincent_analytical_2015}.
%In addition to stochastic switching, an MTJ can also suffer from several reliability issues, the process variation being the main issue. The process variation could severely impact the resistance value of the MTJ and consequently the TMR ratio. Moreover, the STT-MRAM is also prone to other reported reliability challenges including write failure, retention failure and read disturb \cite{fong2013failure} which could lead to errors and potential system failures, depending on the run-time system utilization.

\vspace{-0.5\baselineskip}
\subsection{Designing Spatial-SpinDrop Module}
\label{sec:Designing Spatial-SpinDrop Module}

As mentioned earlier, in the proposed MC-SpatialDropout, feature maps can independently be dropped with a probability $p$. Due to the nature of input application in CiM architectures, this implicitly means dropping different regions of crossbars depending on the mapping strategy. These challenges are associated with designing the Dropout module for the proposed MC-SpatialDropout based BayNN.

% However, there are several challenges associated with designing the Dropout module for the proposed MC-SpatialDropout based BinBayNN. 
For the mapping strategy \textcircled{1}, as depicted in Fig.~\ref{fig:mapping_MWs}(b), each $K\times K$ subset of input comes from a feature map. This means that if an input feature is dropped, the corresponding $K\times K$ subset of input should also be dropped for all $C_{out}$ and $N$ cycles of inputs. This implies that dropping each $K\times K$ row of a crossbar together for $N$ cycles is equivalent to applying spatial dropout. However, each group of rows should be dropped independently of one another. Additionally, their dropout mask should be sampled only in the first cycle. For the remaining $N-1$ cycles of input, the dropout mask should remain consistent.

In contrast, in the mapping strategy \textcircled{2} (see Fig.~\ref{fig:mapping_MWs}(c)), the elements of a MW are applied in parallel to each $K\times K$ crossbar at the same index. As a result, dropping an IF would lead to dropping each index of rows in all the $K\times K$ crossbars together. Similarly, each row of a crossbar is dropped independently of one another, and the dropout mask is sampled at the first input cycle and remains consistent for the remaining $N-1$ cycles of input.

Furthermore, if the spatial dropout is applied to the extracted feature maps of a CNN (see Fig.~\ref{fig:block_drop}), then depending on the usage of the adaptive average pool layer, the design of the Spin-SpatialDrop will differ. If a CNN topology does not use an adaptive average pool layer, then $H\times W$ groups of rows are dropped together. This is because the flattening operation essentially flattens each IF into a vector. These vectors are combined into a larger vector representing the input for the classifier layer. However, since input for the FC layer is applied in one cycle only, there is no need to hold the dropout mask. The Spin-SpatialDrop module for the mapping strategy \textcircled{1} can be adjusted for this condition.

Lastly, if a CNN topology does use an adaptive average pool layer, then the SpinDrop module proposed by~\cite{soyed_nanoarch22,soyed_spindrop} can be used. This is because the adaptive average pool layer averages each IF to a single point, giving a vector with total $C_{out}$ elements.

Therefore, the Dropout module for the proposed MC-SpatialDropout should be able to work in four different configurations. Consequently, we propose a novel spintronic-based spatial Dropout design, called \emph{Spatial-SpinDrop}.

The Spatial-SpinDrop module leverages the stochastic behavior of the MTJ for spatial dropout. The proposed scheme is depicted in Fig.~\ref{fig:dropout_schem}. In order to generate a stochastic bitstream using the MTJ, the first step involves a writing scheme that enables the generation of a bidirectional current through the device. This writing circuit consists of four transistors, allocated to a "SET" and a "RESET" modules. The "SET" operation facilitates the stochastic writing of the MTJ, with a probability corresponding to the required dropout probability. On the other hand, the "RESET" operation restores the MTJ to its original state. 
During the reading operation of the MTJ, the resistance of the device is compared to a reference element to determine its state. The reference resistance value is chosen such as it falls between the parallel and anti-parallel resistances of the MTJ.

For the reading phase, a two-stage architecture is employed for better flexibility and better control of the reading phase for the different configurations discussed earlier. The module operates as follows: after a writing step in the MTJ,  the signal \textit{$V_{pol}$} allows a small current to flow through the MTJ and the reference cell (\textit{REF}), if and only if the signal \textit{$hold$} is activated. Thus, the difference in resistance is translated into a difference in voltages ($V_{MTJ}$ and $V_{ref}$). The second stage of the amplifier utilizes a StrongARM latch structure \cite{razavi_strongarm_2015} to provide a digital representation of the MTJ state. The \textit{Ctrl} signal works in two phases. When \textit{Ctrl = 0},  $\overline{Out}$ and $Out$ are precharged at \textit{VDD}. Later, when \textit{Ctrl = 1}, the discharge begins, resulting in a differential current proportional to the gate voltages ($V_{MTJ}$ and $V_{ref}$). The latch converts the difference of voltage into two opposite logic states in $\overline{Out}$ and $Out$. Once the information from the MTJ is captured and available at the output, the signal \textit{$hold$} is deactivated to anticipate the next writing operation. To enable the dropout, a series of AND gates and transmission gates are added, allowing either access to the classical decoder or to the stochastic word-line (WL).

As long as the \textit{$hold$} signal is deactivated, no further reading operation is permitted. Such a mechanism allows the structure to maintain the same dropout configuration for a given time and will be used during $N-1$ cycles of inputs to allow the dropping of the IF in strategies  \textcircled{1} and \textcircled{2}. In the first strategy, the AND gate receives as input $K\times K$ WLs from the same decoder, see Fig.~\ref{fig:crossbar}(a). While in strategy \textcircled{2}, the AND gate receives one row per decoder, as presented in Fig.~\ref{fig:crossbar}(b). 

For the last two configurations, the  \textit{$hold$} signal is activated for each reading operation, eliminating the need to maintain the dropout mask for $N-1$ cycles.

\vspace{-1\baselineskip}
\subsection{MC-SpatialDropout-Based Bayesian Inference in CiM}
\label{sec:Implementation Spatial-SpinDrop Module to CIM}
% \vspace{-0.7\baselineskip}
%Ones the \textit{$hold$} signal is activated, the \textit{Ctrl} 
% the process involves a comparison between the resistance of the MTJ and a reference resistor.
%The reading operation for MTJs, consists in comparing the resistance of the MTJ with the one
%of a reference, whose value is in between the parallel and anti-parallel resistances. 
%For the reading operation, a two-stages architecture is used for better flexibility. The module operates as follows: after writing in the MTJ,  the signal \textit{$V_{pol}$} allows a small current to flow through the MTJ and a reference cell. 
%The working principle of the module is as follows: the MTJ is initially put in the parallel state, and after a "SET" write operation, the state of the MTJ is read using a sense amplifier. Thereafter, the MTJ is “RESET” to anticipate the next Dropout signal generation.
%the MTJ is initially put in the parallel state, and after a "SET" write operation, the state of the MTJ is read using a sense amplifier to verify if the switching has occurred. Thereafter, the MTJ is “RESET” to its original state to anticipate the next Dropout signal generation.
%Thus, SET and RESET operations are alternated several times to generate a bitstream with a certain probability. Nevertheless, MTJs and CMOS transistors are subject to manufacturing variations that can impact the quality of the generated bitstream and introduce a deviation from the target probability.
\begin{figure}
\centering
   \subfloat[]{\includegraphics[width=0.25\columnwidth, trim={0 -3cm 0 0}]{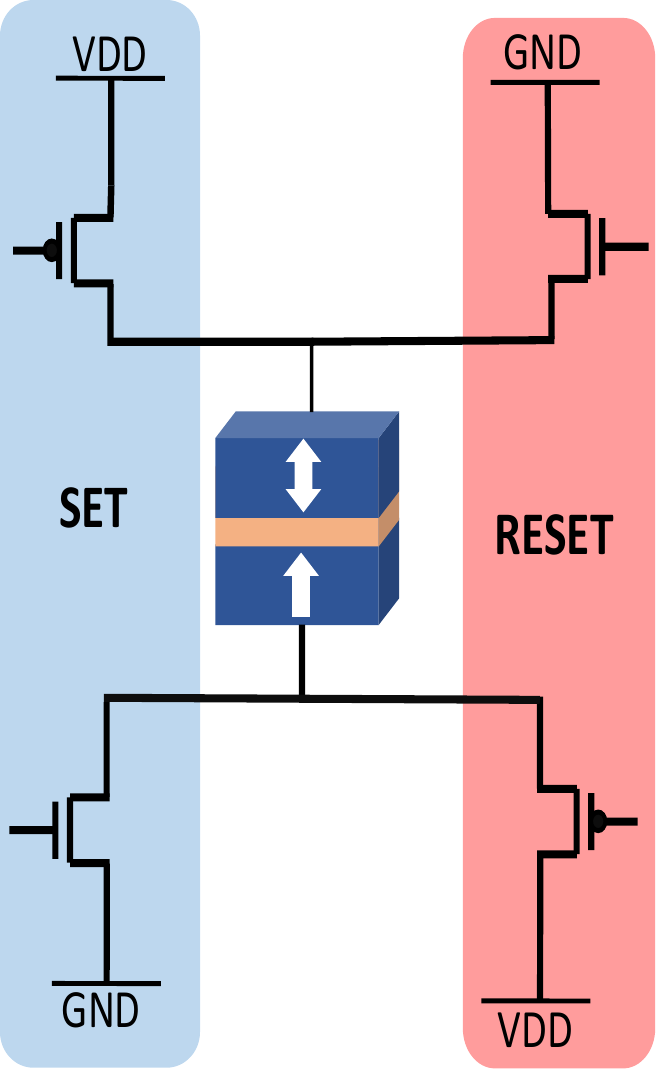}}
    \hspace{1em}
    \subfloat[]{\includegraphics[width=0.65\columnwidth]{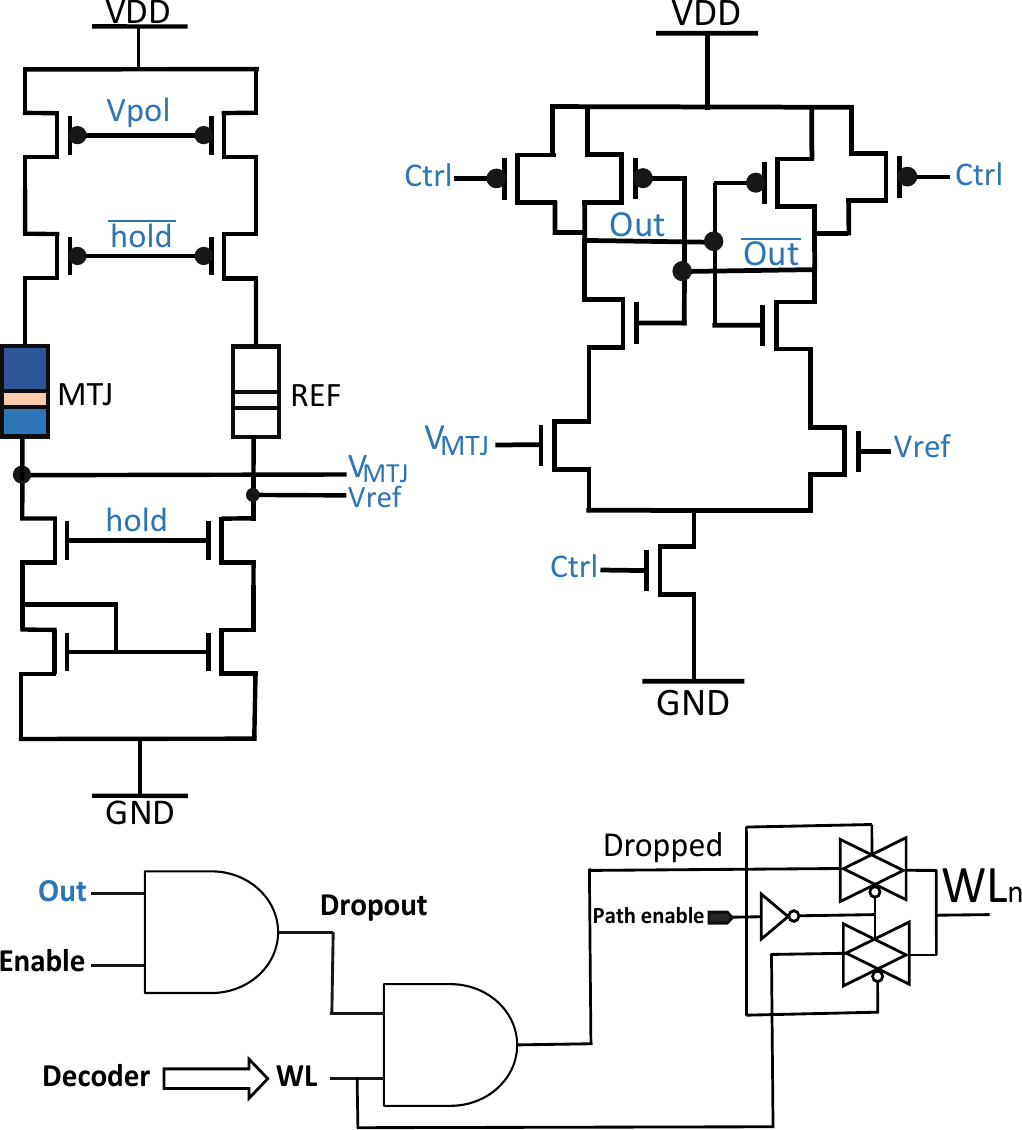}}
  \caption{(a) Writing and (b) reading schemes for the MTJ.}
\label{fig:dropout_schem}
\end{figure}
% I NEED TO ADD FIGURE difference(DROPOUT).
The proposed MC-SpatialDropout-Based Bayesian inference can be leveraged on the two mapping strategies discussed in Section~\ref{sec:mapping}. In both strategies, one or more crossbar arrays with MTJs at each crosspoints are employed in order to encode the binary weights into the resistive states of the MTJs.
% For the first strategy, each 3D kernel is unrolled into a column of a crossbar with a size of $K\times K\times C_{in}$ \cite{gokmen_training_2017}. In such a strategy, 
\begin{figure}
\centering
   \includegraphics[width=0.9\columnwidth]{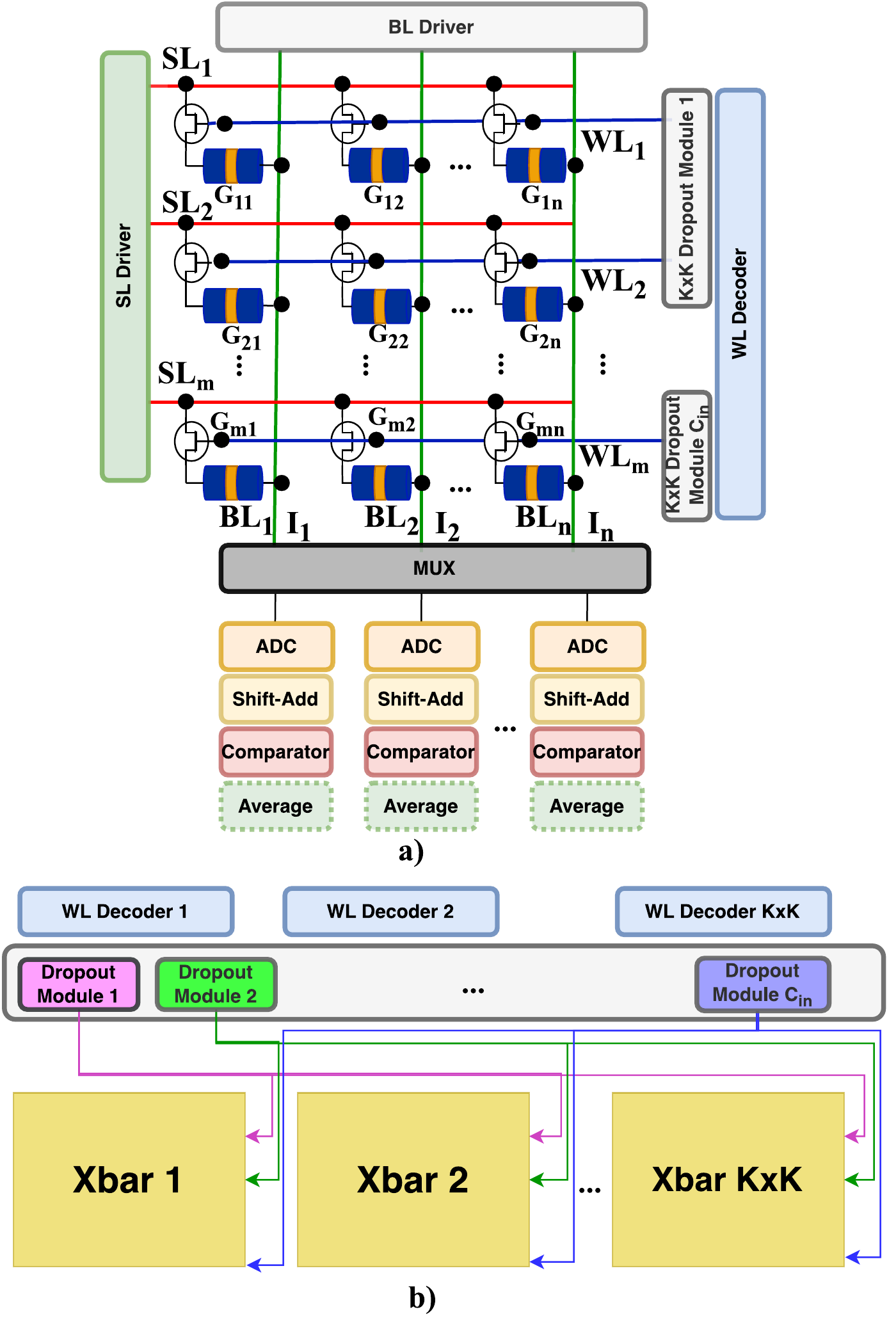}
\vspace{-0.5\baselineskip}   
  \caption{Crossbar design for the MC-SpatialDropout based on mapping strategy (a)  \textcircled{1} and (b) strategy\textcircled{2}. In (b), only the Dropout module and WL decoder are shown, Everything else is abstracted. }
\label{fig:crossbar}
\end{figure}

Specifically, for the mapping strategy \textcircled{1}, we divide the WLs of the crossbar into $K\times K$ groups and connect one dropout module to each group, as shown in Fig.\ref{fig:crossbar}(a). In Fig.~\ref{fig:dropout_schem}(b), this strategy involves connecting $K\times K$ WLs to an AND gate. The AND gate receives the signal delivered by the decoder as its input. This configuration allows for the selective activation or deactivation of a group of WLs. To facilitate the activation of multiple consecutive addresses in the array, an adapted WL decoder is utilized.
% A dropout module is connected to $K\times K$ WLs for the spatial dropout.
The bit-line and source-line drivers were used to manage the analog input and output for the MVM operation. Also, a group-wise selection of WLs is performed concurrently and the intermediate result for MVM operation is accumulated into an accumulator block until all the WLs are selected for each layer. We utilized MUXes to select the different bit-lines that are sensed and converted by ADC.  The shift-adder modules are used to shift and accumulate the partial sums coming from the array. Finally, a digital comparator and averaging block are used to implement the activation function. For the last layer, the average operation is performed with an averaging block. 

For the mapping strategy \textcircled{2}, a similar architecture to strategy \textcircled{1} is employed. The key distinction relies upon the utilization of $K\times K$ crossbars in parallel to map the binary weights of a layer. Also, the dropout modules are connected to a similar WL index in each of the crossbar arrays, as shown in Fig.~\ref{fig:crossbar}(b). Here, the same AND gate in the Dropout module receives signals from different decoders and the result is sent to each row of the $K\times K$ crossbars. For instance, the first WL of each crossbar of a layer connects the same Dropout module. All the WLs decoders are connected to a dropout block in gray in Fig.~\ref{fig:crossbar}(b) comprising $C_{in}$ dropout modules. It is worth mentioning that the dropout is used during the reading phase only, therefore, the dropout module is deactivated during the writing operation and WL decoders are used normally. 
% \cite{peng_optimizing_2019} consists in unrolling weights at various spatial positions within each kernel. Thus,  $K\times K$ submatrices are used instead of a large matrix. In this second configuration,

%In terms of crossbar mapping of the convolutional layer with the dropout,
% \begin{figure}
%     \centering
%     \includegraphics[width=\columnwidth]{figures/mappings_updated.pdf}
%     \caption{a) Output feature maps of a Convolution layer. Input processing for the mapping strategies b) \textcircled{1}, and c) \textcircled{2} with respective (implicit) row dropout. In \textcircled{1}, a $K\times K$ group of rows "drop" together in a crossbar. In \textcircled{2}, same index of each row drop together.}
%     \label{fig:mapping}
% \end{figure}

%\begin{figure}
%\centering
  % \includegraphics[width=1\columnwidth]{figures/xbar3.pdf}
%\label{fig:dropout_schem}
%\end{figure}

%\begin{figure}
%\centering
%   \includegraphics[width=0.9\columnwidth]{figures/design6.pdf}
%   \caption{(a) Crossbar design based on strategy \textcircled{1} and (b)strategy\textcircled{2} }
%\label{fig:crossbar}
%\end{figure}

% \begin{figure}
% \centering
%    \subfloat[]{\includegraphics[width=0.8\columnwidth]{figures/Drop_crossbar.pdf}}
%     \hspace{0.1em}
%     \subfloat[]{\includegraphics[width=0.5\columnwidth]{figures/machin2.pdf}}
%   \caption{(a) Crossbar design based on strategy \textcircled{1} and (b) strategy\textcircled{2} }
% \label{fig:crossbar}
% \end{figure}

%\begin{figure}
%\centering
 %  \includegraphics[width=1.0\columnwidth]{figures/x2-1.png}
%\label{fig:dropout_schem}
%\end{figure}

\section{Results}
\label{sec:Results}
\subsection{Simulation Setup}
\label{sec:Simulation Setup}
% Note that the scalability of the proposed method is not an issue as 

We evaluated the proposed MC-SpatialDropout on predictive performance using VGG, ResNet-18, and ResNet-20 topologies on the CIFAR-10 dataset. All the models were trained with SGD optimization algorithm, minimizing the proposed learning objective~\eqref{eq:MC-SpatialDropout} with $\lambda$ chosen between $1\times 10^{-5}$ and $1\times 10^{-7}$, and the binarization algorithm from \cite{irnet} was used. Also, all the models are trained with $\rho=15\%$ dropout probability. The validation dataset of the CIFAR-10 is split 80:20 with 20\% of the data used for the cross-validation and 80\% used for evaluation. 

% As our work integrates  established methods, e.g., the BNN algorithm from \cite{}, and the dropout method from the SpatialDropout paper. By integrating these established methods, our approach inherits their scalability to larger datasets, as demonstrated in previous studies. Notably, these techniques have been successfully applied to datasets like ImageNet, showcasing the potential scalability of our approach to handle larger-scale datasets effectively.

To assess the effectiveness of our method in handling uncertainty, we generated six additional OOD datasets:
\begin{enumerate*}
  \item Gaussian noise ($\hat{\mathcal{D}}_1$): Each pixel of the image is generated by sampling random noise from a unit Gaussian distribution, $\x\sim \mathcal{N}(0,1)$,
  \item Uniform noise ($\hat{\mathcal{D}}_2$): Each pixel of the image is generated by sampling random noise from a uniform distribution,  $\x \sim \mathcal{U}(0,1)$,
  \item CIFAR-10 with Gaussian noise ($\hat{\mathcal{D}}_3$): Each pixel of the CIFAR-10 images is corrupted with Gaussian noise,
  \item CIFAR-10 with uniform noise ($\hat{\mathcal{D}}_4$): Each pixel of the CIFAR-10 images is corrupted with uniform noise,
  \item SVHN: Google street view house numbers dataset, and
  \item STL10: a dataset containing images from the popular ImageNet dataset.
\end{enumerate*}
Each of these OOD datasets contains $8000$ images, and the images have the same dimensions as the original CIFAR-10 dataset ($32 \times 32$ pixels). During the evaluation phase, 
% we assessed the uncertainty estimation performance of all models utilizing uncertainty estimation metrics such as 
% To classify 
an input is classified as OOD or ID as follows:
% \begin{algorithm}
% \caption{Uncertainty Estimation and Prediction Acceptance/Rejection}
% \begin{algorithmic}[1]
% \REQUIRE Input image $X$
% \REQUIRE Trained model
% \REQUIRE Number of stochastic forward passes $T$
% \STATE Initialize $R = 0$, $A = 0$ \COMMENT{Flags for rejected and accepted prediction}
% \FOR{$i = 1$ to $T$}
%     \STATE Perform stochastic forward pass with the model
%     \STATE Record softmax output $Y_{i}$
% \ENDFOR
% \STATE Calculate the 10th percentile (quantile 0.1) across $Y_{1}, Y_{2}, \ldots, Y_{T}$ at the neuron dimension to get a probability distribution across all classes
% \STATE Determine the maximum probability $p = \max(\text{quantile}_{0.1}(Y_{1}, Y_{2}, \ldots, Y_{T}))$
% \IF{$p < 0.9$}
%     \STATE $R = 1$ \COMMENT{Prediction is rejected due to high uncertainty}
% \ELSE
%     \STATE $A = 1$ \COMMENT{Prediction is accepted as the model is estimated to be certain and uncertainty is low}
% \ENDIF
% \RETURN $R$, $A$
% \end{algorithmic}
% \end{algorithm}

\begin{equation}
\begin{cases} 
\text{OOD}, & \text{if } \max\left(\mathcal{Q}\left(\frac{1}{T}\sum_{t=1}^{T} \y_t\right)\right) < 0.9 \\
\text{ID}, & \text{otherwise}.
\end{cases}
\end{equation}
Here, $\y_t$ is the softmax output of the stochastic forward pass at MC run $t$ with $T$ MC runs, the function $\mathcal{Q}(\cdot)$ calculates the 10th percentile across a set of values, and the function $\max(\cdot)$  determines the maximum confidence score across output classes. Overall, OOD or ID is determined by whether the maximum value from the 10th percentile of the averaged outputs is less than 0.9 (for OOD) or not (for ID). The intuition behind our OOD detection is that the majority of confidence score of the $T$ MC runs is expected to be high and close to one another (low variance) for ID data and vice versa for OOD data.

The hardware-level simulations for the proposed method were conducted on the Cadence Virtuoso simulator with 28nm-FDSOI STMicroelectronics technology library for the respective network topologies and dataset configurations. 
%The simulator was configured to operate with the xx processing unit and xxx memory architecture.

\vspace{-0.5\baselineskip}
\subsection{Predictive Performance and Uncertainty Estimation}
\label{sec:Predictive Performance}
\begin{table}
\centering
\caption{Predictive Performance of the proposed MC-SparialDropout method in comparison with SOTA methods on CIFAR-10.}
\resizebox{\linewidth}{!}{
\begin{tabular}{|c|c|c|c|c|}
\hline
Topology                   & Method            & Bit-width (W/A) & Bayesian & Inference Accuracy \\ \hline
\multirow{5}{*}{ResNet-18} & FP                & 32/32           &       No   & $93.0\%$           \\ \cline{2-5} 
                           & RAD~\cite{rad}               & 1/1             &     No     & $90.5\%$           \\ \cline{2-5} 
                           & IR-Net~\cite{irnet}            & 1/1             &     No     & $91.5\%$           \\ \cline{2-5} 
                           & SpinDrop~\cite{soyed_nanoarch22,soyed_spindrop}          & 1/1             &     Yes     & $90.48\%$              \\ \cline{2-5} 
                           & \textbf{Proposed} & 1/1             &     Yes     & $\mathbf{91.34\%}$              \\ \hline
\multirow{6}{*}{ResNet-20} & FP                & 32/32           &       No   & $91.7\%$           \\ \cline{2-5} 
                           & DoReFa~\cite{dorefa}            & 1/1             &     No     & $79.3\%$           \\ \cline{2-5} 
                           % & LQ-Net            & 1/1             &     No     & $90\%$           \\ \cline{2-5} 
                           & DSQ~\cite{dsq}               & 1/1             &     No     & $84.1\%$           \\ \cline{2-5} 
                           & IR-Net~\cite{irnet}            & 1/1             &     No     & $85.4\%$           \\ \cline{2-5} 
                           & \textbf{Proposed} & 1/1             &     Yes     & $\mathbf{84.71\%}$            \\ \hline
\multirow{8}{*}{VGG}       & FP                & 32/32           &       No   & $91.7\%$           \\ \cline{2-5} 
                           & LAB~\cite{LAB}               & 1/1             &     No     & $87.7\%$           \\ \cline{2-5} 
                           & XNOR~\cite{xnor}              & 1/1             &     No     & $89.8\%$           \\ \cline{2-5} 
                           & BNN~\cite{bnn}               & 1/1             &     No     & $89.9\%$           \\ \cline{2-5} 
                           & RAD~\cite{rad}               & 1/1             &     No     & $90.0\%$           \\ \cline{2-5} 
                           & IR-Net~\cite{irnet}            & 1/1             &     No     & $90.4\%$           \\ \cline{2-5} 
                           & SpinDrop~\cite{soyed_nanoarch22,soyed_spindrop}          & 1/1             &     Yes     & $91.95\%$            \\ \cline{2-5} 
                           & \textbf{Proposed} & 1/1             &     Yes     & $\mathbf{90.34\%}$            \\ \hline
\end{tabular}
}
\label{tab:pred_per}
\end{table}

The predictive performance of the approach is close to the existing conventional BNNs, as shown in Table~\ref{tab:pred_per}. Furthermore, in comparison to Bayesian approaches~\cite{soyed_nanoarch22,soyed_spindrop}, our proposed approach is within $1\%$ accuracy. Furthermore, the application of Spatial-SpinDrop before the convolutional layer and at the extracted feature maps can also achieve comparable performance ($\sim 0.2\%$), see Fig.~\ref{fig:block_drop}. This demonstrates the capability of the proposed approach in achieving high predictive performance. However, note that applying Spatial-SpinDrop before all the convolutional layers can reduce the performance drastically, e.g., accuracy reduces to $75\%$ on VGG. This is because at shallower layers, the number of OFMs is lower in comparison, leading to a high chance that most of the OFMs are being omitted (dropped). Also, as shown by~\cite{soyed_nanoarch22,soyed_spindrop}, BNNs are more sensitive to the dropout rate. Therefore, a lower Dropout probability between $10-20\%$ is suggested.

In terms of OOD detection, our proposed method can achieve up to 100\% OOD detection rate across various model architectures and six different OOD datasets ($\hat{\mathcal{D}}_1$ through $\hat{\mathcal{D}}_6$), as depicted in Table~\ref{tab:ood}. There are some variations across different architectures and OOD datasets. 
% with ResNet-20 and VGG showing slightly lower detection rates on $\hat{\mathcal{D}}_3$ and $\hat{\mathcal{D}}_4$ datasets. 
However, even in these cases, our method can consistently achieve a high OOD detection rate, with the lowest detection rate being $64.39\%$ on the ResNet-18 model with $\hat{\mathcal{D}}_4$ dataset and Spatial-SpinDrop applied to extracted feature maps. However, when the Spatial-SpinDrop is applied to the convolutional layers of the last residual block, OOD detection rate on $\hat{\mathcal{D}}_4$ dataset improved to $97.39\%$, a  $33.00\%$ improvement. Therefore, we suggest applying the Spatial-SpinDrop to the last convolutional layers to achieve a higher OOD detection rate at the cost of a small accuracy reduction. Consequently, the result suggests that the MC-SpatialDropout method is a robust and reliable approach to OOD detection across various model architectures and datasets.

% \subsection{Uncertainty Estimation}
% \label{sec:Uncertainty Estimation}
\begin{table}
\caption{Evaluation of the proposed MC-SpatialDropout method in detecting OOD.}
\begin{threeparttable}
\resizebox{\linewidth}{!}{
\begin{tabular}{|c|c|c|c|c|c|c|}
\hline
Topologie     & $\hat{\mathcal{D}}_1$     & $\hat{\mathcal{D}}_2$   & $\hat{\mathcal{D}}_3$     & $\hat{\mathcal{D}}_4$     & $\hat{\mathcal{D}}_5$     & $\hat{\mathcal{D}}_6$     \\ \hline
ResNet-18     & $99.56\%$   & $99.94\%$ & $96.1\%$   & $81.68\%$   & $83.02\%$   & $64.39\%$   \\ \hline
ResNet-18\tnote{i} & $100\%$   & $100\%$ & $100\%$   & $92.26\%$   & $99.98\%$   & $97.39\%$   \\ \hline
ResNet-20     & $97.2\%$  & $100\%$ & $90.79\%$ & $87.94\%$ & $99.03\%$ & $99.81\%$ \\ \hline
VGG           & $99.99\%$ & $100\%$ & $92.9\%$  & $78.91\%$ & $99.81\%$ & $100\%$   \\ \hline
\end{tabular}
}
\begin{tablenotes}
\item[i] Spatial-Dropout applied to final two convolutional layers.
\end{tablenotes}
\end{threeparttable}
\label{tab:ood}
\end{table}

\vspace{-1\baselineskip}
\subsection{Overhead Analysis}
\label{sec:Overhead Analysis}
The proposed Spatial-SpinDrop modules were evaluated for the area, power consumption, and latency as shown in Table~\ref{tab:drop_modules} and compared with the SpinDrop approach presented in ~\cite{soyed_nanoarch22,soyed_spindrop}. These evaluations were conducted using a crossbar array with dimensions of $64\times32$ and scaled for the VGG topology. In layer-wise application of spatial Dropout, the Dropout modules applied to convolutional layers of the last VGG block. Also, for topology-wise application of spatial Dropout, Dropout modules are applied to the extracted feature maps. In our evaluation, a configuration of $C_{in}=256$, $K=3$ and $C_{out}=512$ is used.

At first, in terms of area, the SpinDrop method requires one dropout module per row in the crossbar structure, while our method only requires one dropout module per $K\times K$ group of rows. Therefore, the area and the power consumption of dropout modules are reduced by a factor of $9$. In terms of latency for the dropout modules, we achieve  $15 ns$ in all cases. Indeed, to generate 1 bit, for a given number of rows, the dropout module needs to be written, however, such latency can be further decreased by increasing the writing voltages of the MTJ. Furthermore, in the case, the adaptive average pool layer is not used, the power consumption and the area for the SpinDrop approach increases greatly ($\times 9$). While in the proposed approach, the adaptative Average pool layer does not impact the total energy and area, as mentioned in Section~\ref{sec:Designing Spatial-SpinDrop Module} and shown in Table~\ref{tab:drop_modules}.

Table~\ref{tab:energy_sota} compares the energy consumption of the proposed approach with the State-Of-The-Art implementation based on the MNIST dataset. For the evaluation, we used NVSIM, and we estimated the total energy for a LeNet-5 architecture to be consistent with the approach presented in~\cite{soyed_nanoarch22}. When compared to the SpinDrop approach in~\cite{soyed_nanoarch22} our approach is $2.94\times$ more energy efficient. Furthermore, when compared to RRAM technology, our solution is $13.67\times$ more efficient. Finally, in a comparison with classic FPGA implementation, the proposed approach achieves substantial energy savings of up to $94.11\times $.

\begin{table}
\vspace{-1\baselineskip}
\centering
\caption{Layer-wise Overhead Analysis of the Proposed Method in Comparison to SpinDrop~\cite{soyed_nanoarch22,soyed_spindrop}.}
\resizebox{\linewidth}{!}{
\begin{tabular}{|cccccc|}
\hline
\multicolumn{6}{|c|}{Layer-wise application of spatial Dropout}                                                                                                                                                                                                                                                                                                                                                                                    \\ \hline
\multicolumn{1}{|c|}{Method}                             & \multicolumn{1}{c|}{\begin{tabular}[c]{@{}c@{}}Mapping \\ Strategy\end{tabular}}   & \multicolumn{1}{c|}{\begin{tabular}[c]{@{}c@{}}\# of Dropout \\ Modules\end{tabular}} & \multicolumn{1}{c|}{Area} & \multicolumn{1}{c|}{\begin{tabular}[c]{@{}c@{}}Power\\ Consumption\end{tabular}} & \begin{tabular}[c]{@{}c@{}}Sampling \\ Latency\end{tabular} \\ \hline
\multicolumn{1}{|c|}{\multirow{2}{*}{SpinDrop}}          & \multicolumn{1}{c|}{\textcircled{1}}                                                             & \multicolumn{1}{c|}{$K*K*C_{in}$}                                                     & \multicolumn{1}{c|}{$79833.6\mu m^2$}     & \multicolumn{1}{c|}{$51.84 mW$}                                                             & $15 ns$                                                            \\ \cline{2-6} 
\multicolumn{1}{|c|}{}                                   & \multicolumn{1}{c|}{\textcircled{2}}                                                             & \multicolumn{1}{c|}{$K*K*C_{in}$}                                                     & \multicolumn{1}{c|}{$79833.6\mu m^2$}     & \multicolumn{1}{c|}{$51.84 mW$}                                                             &   $15 ns$                                                           \\ \hline
\multicolumn{1}{|c|}{\multirow{2}{*}{\textbf{Proposed}}} & \multicolumn{1}{c|}{\textcircled{1}}                                                             & \multicolumn{1}{c|}{$C_{in}$}                                                         & \multicolumn{1}{c|}{$8870.4 \mu m^2$}     & \multicolumn{1}{c|}{$5.76 mW$}                                                             &       $15 ns$                                                       \\ \cline{2-6} 
\multicolumn{1}{|c|}{}                                   & \multicolumn{1}{c|}{\textcircled{2}}                                                             & \multicolumn{1}{c|}{$C_{in}$}                                                         & \multicolumn{1}{c|}{$8870.4 \mu m^2$}     & \multicolumn{1}{c|}{$5.76 mW$}                                                             &       $15 ns$                                                      \\ \hline
\multicolumn{6}{|c|}{Topology-wise application of spatial Dropout}                                                                                                                                                                                                                                                                                                                                                                                 \\ \hline
\multicolumn{1}{|c|}{Method}                             & \multicolumn{1}{c|}{\begin{tabular}[c]{@{}c@{}}Adaptive \\ Avg. Pool\end{tabular}} & \multicolumn{1}{c|}{\begin{tabular}[c]{@{}c@{}}\# of Dropout\\ Modules\end{tabular}}  & \multicolumn{1}{c|}{Area} & \multicolumn{1}{c|}{\begin{tabular}[c]{@{}c@{}}Power\\ Consumption\end{tabular}}  & \begin{tabular}[c]{@{}c@{}}Sampling\\ Latency\end{tabular}  \\ \hline
\multicolumn{1}{|c|}{\multirow{2}{*}{SpinDrop}}          & \multicolumn{1}{c|}{Used}                                                          & \multicolumn{1}{c|}{$C_{out}$}                                                         & \multicolumn{1}{c|}{$17740.8 \mu m^2$}     & \multicolumn{1}{c|}{$11.52 mW$}                                                      &    $15 ns$                                                        \\ \cline{2-6} 
\multicolumn{1}{|c|}{}                                   & \multicolumn{1}{c|}{Not Used}                                                      & \multicolumn{1}{c|}{$K*K*C_{out}$}                                                     & \multicolumn{1}{c|}{$159 667.2\mu m^2$}     & \multicolumn{1}{c|}{$103.68 mW$}                                                             &                                        $15 ns$                      \\ \hline
\multicolumn{1}{|c|}{\multirow{2}{*}{\textbf{Proposed}}}          & \multicolumn{1}{c|}{Used}                                                          & \multicolumn{1}{c|}{$C_{out}$}                                                         & \multicolumn{1}{c|}{$17740.8 \mu m^2$}     & \multicolumn{1}{c|}{$11.52 mW$}                                                             & $15 ns$                                                            \\ \cline{2-6} 
\multicolumn{1}{|c|}{}                                   & \multicolumn{1}{c|}{Not Used}                                                      & \multicolumn{1}{c|}{$C_{out}$}                                                     & \multicolumn{1}{c|}{$17740.8 \mu m^2$}     & \multicolumn{1}{c|}{$11.52 mW$}                                                             &                                                       $15 ns$       \\ \hline
\end{tabular}}

\label{tab:drop_modules}
\end{table}

\begin{table}[hbt]
\caption{Energy Efficiency Comparison of Hardware Implementations}
\resizebox{\columnwidth}{!}
{
\centering
\begin{tabular}{|l|l|l|l|}
\hline
Related works     & Technology & Bit resolution & Energy \\ \hline
R.Cai et al.\cite{cai_vibnn_2018}      & FPGA  & 8-bit         & 18.97~\SI{}{\micro\joule}/Image         \\ \hline
X.Jia et al.\cite{jia_efficient_2021}      & FPGA  & 8-bit        & 46.00~\SI{}{\micro\joule}/Image  \\ \hline
H.Awano et al. \cite{awano_bynqnet_2020}    & FPGA      & 7-bit     & 21.09~\SI{}{\micro\joule}/Image           \\ \hline
A. Malhotra et al. \cite{malhotra_exploiting_2020}      & RRAM  & 4-bit        & 9.30~\SI{}{\micro\joule}/Image         \\ \hline
S.T.Ahmed et al.\cite{soyed_nanoarch22} & STT-MRAM   & 1-bit     & 2.00~\SI{}{\micro\joule}/Image              \\ \hline
\textbf{Proposed implementation} & \textbf{STT-MRAM}   & \textbf{1-bit}     & \textbf{0.68}~\SI{}{\micro\joule}/Image  \\ \hline
\end{tabular}
}
\label{tab:energy_sota}
\end{table}

\vspace{-0.5\baselineskip}
\section{Conclusion}\label{sec:conclusion}
In this paper, we present MC-SpatialDropout, an efficient spatial dropout-based approximation for Bayesian neural networks. The proposed method exploits the probabilistic nature of spintronic technology to enable Bayesian inference. Implemented on a spintronic-based Computation-in-Memory fabric with STT-MRAM, MC-SpatialDropout achieves improved computational efficiency and power consumption.

\section*{Acknowledgments}
This work was supported by a joint ANR-DFG grant Neuspin Project ANR-21-FAI1-0008. 

\bibliographystyle{IEEEtran}

\bibliography{references}


% \end{IEEEbiography}


\vfill

\end{document}